\documentclass[11pt]{article}
\usepackage{nodalida2025}
\usepackage{enumitem}
\usepackage{times}
\usepackage{url}
\usepackage{latexsym}
\usepackage{pgfplots}
\pgfplotsset{width=6.5cm, compat=1.17}
\usepackage{subcaption} 
\usepackage{url}
\usepackage{hyperref}
\hypersetup{breaklinks=true}
\usepackage[hyphenbreaks]{breakurl}

\aclfinalcopy 

\title{BiaSWE: An Expert Annotated Dataset for Misogyny Detection in Swedish}

\author{
  Kätriin Kukk \textsuperscript{1,2}, Danila Petrelli\textsuperscript{1}, Judit Casademont\textsuperscript{1},\\ \textbf{Eric J. W. Orlowski\textsuperscript{3},}  \textbf{Michał Dzieliński\textsuperscript{4},} \textbf{Maria Jacobson\textsuperscript{5}}\\
  \textsuperscript{1}AI Sweden\\
  \textsuperscript{2}Linköping University\\
  \textsuperscript{3}AI Singapore\\
  \textsuperscript{4}Stockholm University\\
  \textsuperscript{5}Anti-Discrimination Agency West Sweden\\
  {\tt katriin.kukk@liu.se}, {\tt danila.petrelli@ai.se}, \\
  {\tt juditcasademont@gmail.com}, {\tt ericorlowski@aisingapore.org}, \\
  {\tt michal.dzielinski@sbs.su.se}, {\tt maria.jacobson@adbvast.se}
}

\date{}

\begin{document}
\maketitle
\begin{abstract}
  In this study, we introduce the process for creating BiaSWE, an expert-annotated dataset tailored for misogyny detection in the Swedish language. To address the cultural and linguistic specificity of misogyny in Swedish, we collaborated with experts from the social sciences and humanities. Our interdisciplinary team developed a rigorous annotation process, incorporating both domain knowledge and language expertise, to capture the nuances of misogyny in a Swedish context. This methodology ensures that the dataset is not only culturally relevant but also aligned with broader efforts in bias detection for low-resource languages. The dataset, along with the annotation guidelines, is publicly available for further research.
  \end{abstract}

\section{Introduction}
\label{Introduction}
Large Language Models (LLMs) have experienced immense growth over the past years due to being capable of solving diverse tasks that previously required a separate model for each specific task \citep{de_angelis_et_al_2023}. Despite their apparent benefits, it is known that the characteristics of the dataset used to train a language model play a fundamental role in determining the model’s behavior \citep{gebru_datasheets}. LLMs are typically trained on large amounts of data from the Internet and thus inevitably reflect the opinions and biases of its users. For example, a 2018 survey showed that about 85\% of English Wikipedia contributors identified as male \citep{oldach2022wikipedia}. As LLMs' behavior ``reflects the Collective Intelligence of Western society'', LLMs can perpetuate and even amplify biases and stereotypes of social minorities \citep{kotek_et_al_2023}. The widespread presence of misogyny online is illustrated by a study from 2020 where 65\% of women reported knowing another woman that had been the target of online violence \citep{eiu2020infographic}.

The way to avoid harmful machine learning models is to ensure that the datasets used for training are responsibly curated, involving diverse stakeholders \citep{delgado2021stakeholder}. However, dataset creation alone is not sufficient, and additional approaches, such as alignment, play a role in guiding model outputs towards human values. In the context of bias detection, misogyny varies by language and culture \citep{zeinert-etal-2021-annotating}. Therefore, we consider creating expert-annotated, language-specific datasets crucial for detecting biases, helping to identify areas where models may risk perpetuating harmful stereotypes or undesirable attitudes.

To address these challenges, we make two key contributions\textsuperscript{1}
\footnotetext[1]{\raggedright Link to the dataset and annotation guidelines:
https://huggingface.co/datasets/AI-Sweden-Models/BiaSWE \label{footnote:guidelines}}:\\ 
1. We present BiaSWE, a small annotated dataset for misogyny detection in Swedish, annotated for hate speech, misogyny, misogyny type categories and severity.\\
2. We share the creation process of the BiaSWE dataset and our annotation guidelines. By doing this, we show how an existing experiment can be adapted to the Swedish needs and cultural context.

\section{Related Work}
In recent years, work has been done in the field of dataset creation for bias and hate speech in general, paying great attention to data coming from online sources, especially social media, such as Twitter, Facebook, Reddit, or blogs. This kind of work has been carried out in a multitude of languages, across several cultural contexts, and tends to cover various forms of sexism as it presents in written language. This is the case of \citet{chiril-etal-2020-annotated}, who present a corpus for detecting sexism in French tweets. Another example is the work of \citet{zeinert-etal-2021-annotating}, who sample their Bajer dataset from Twitter, Facebook and Reddit posts in Danish. However, research is also carried out with the target of more subtle, less explicit misogyny in mind; this is the case of the Biasly dataset by \citet{sheppard-etal-2024-biasly}, who gathered their data from scripts from North American movies, in English.

The most common method for data collection among the different existing datasets is using keywords \citep{chiril-etal-2020-annotated, zeinert-etal-2021-annotating, sheppard-etal-2024-biasly}. The degree of detail or the number of keywords varies from words that do not necessarily imply misogyny (e.g. ``she'') to ambiguous keywords, to keywords that are very highly related to misogyny and sexism (e.g. ``\#MeToo'').

Most of the existing misogyny detection datasets provide a taxonomy for different categories of misogyny in addition to the binary classification. Many regard the addition of a multi-label classification layer as necessary, given that ``binary detection […] disregards the diversity of sexist content, and fails to provide clear explanations for why something is sexist'' \citep{kirk-etal-2023-semeval}. There is no clear consensus regarding the types of misogyny to classify the sentences into, or even on the optimal level of detail regarding the categories.

To the best of our knowledge, this work is the first attempt to create resources for misogyny detection for the Swedish language.

\section{Method}

This section provides an overview of our data preparation process, introduces the team of expert annotators and details the annotation workflow.

\subsection{Data}
As a data source for our dataset, we used the Swedish website Flashback, one of the largest Swedish internet forums since the 1990s. Known for its focus on freedom of speech, the forum hosts discussions on controversial subjects, and its anonymity often leads to misuse \citep{norlund-stenbom-2021-building}.

To ensure the presence of enough misogynistic examples in the final dataset, we decided to use keyword search. Taking into account the cultural and linguistic closeness of Danish and Swedish, our initial list of keywords was based on the work of \citet{zeinert-etal-2021-annotating} that used both keywords and hashtags in Danish. We excluded all hashtags but ``\#MeToo'' because of their rarity on Flashback. The keywords were first translated from Danish into Swedish with the help of a Danish speaker. Thereafter, we presented the resulting Swedish keywords to our team of expert annotators, who suggested removing some of the keywords and adding others. Our list consists of 118 key terms including words and phrases in Swedish (e.g., ``kvinna'') as well as some terms and names in popular English slang (e.g., ``Chad''). The full list of keywords is available in the annotation guidelines (see section \ref{Introduction}).

Based on these keywords, we gathered 450 data points, each to be annotated by two or more annotators. We did not want several annotators to have the exact same set of data points, so we used a rotation system that distributed them. Each expert was assigned 210 data points.

\subsection{Annotation}
The team of annotators included researchers and experts from the humanities and social sciences, as well as civil society actors. Four of our experts identify as women and the other three as men. Our experts volunteered to participate in the project amongst a bigger pool of experts in humanities, social sciences, and civil society representatives that have been introduced to the basics of LLMs and AI within a broader interdisciplinary project at AI Sweden\textsuperscript{2}
\footnotetext[2]{\raggedright Link to the project page: https://www.ai.se/en/project/interdisciplinary-expert-pool-nlu}. From this point on, we refer to them as 'experts', acknowledging their role in both annotation and providing critical insights into the interdisciplinary process.

To facilitate the annotation process, we prepared annotation guidelines (see section \ref{Introduction}) by taking inspiration from the work by \citet{sheppard-etal-2024-biasly}. Their guidelines included a definition of misogyny that was modified together with the team of experts to obtain the following final definition\textsuperscript{3}:
\begin{quote}
Hatred of, dislike of, contempt for, ingrained prejudice, control of or oppression against women as well as going against the idea of feminism. It is a form of sexism and can be either intentional or unintentional. Misogyny can contain different types of opinions and values such as seeing women as inferior, asserting men's sexual entitlement, objectifying women, accepting violence, celebrating traditional gender roles but also the need to "protect" women as well as thinking that equality and feminism have gone too far.
One of the ways misogyny can be expressed is through language and, in this project, we focus on misogynistic language portrayed in text. Misogyny can be perpetrated by people regardless of their gender.\end{quote}
\footnotetext[3]{Link to the complete guidelines, as well as the definition in Swedish, can be found on Hugging Face (see Footnote~\ref{footnote:guidelines}).}

We also present a taxonomy of five misogyny categories constructed by combining the twelve categories by \citet{sheppard-etal-2024-biasly} and the six categories by \citet{zeinert-etal-2021-annotating} and modifying these with the help of the team of experts. The guidelines also give detailed instructions on how to carry out the annotation task and provide examples. For annotation, we used an open-source platform called Label Studio. After everyone had annotated up to 50 examples, we held a workshop to discuss examples where the experts had disagreed.

We divided the annotation of each data point into four small tasks, each more fine-grained than the previous. Once a negative answer was given by the expert, the annotation process ended and the downstream tasks did not need to be completed.

\begin{figure}
    \centering
    \begin{tikzpicture}
        \begin{axis}[
                ybar,
                symbolic x coords={A, B, C, D},
                xtick=data,
                ylabel={Number of posts},
                xticklabel style={rotate=90, anchor=east, align=center},
                xticklabels={
                    {Annotated\\posts}, 
                    {Yes,\\all annotators}, 
                    {No,\\all annotators}, 
                    {Yes, at least\\one annotator}
                },
                ymin=0,
                bar width=15pt,
                nodes near coords,
                enlarge x limits=0.2,
                enlarge y limits={upper, value=0.2}
                ]
        \addplot coordinates {(A, 450)
                             (B, 139) 
                             (C, 207)
                             (D, 243)
                             };
        \end{axis}
    \end{tikzpicture}
    \caption{Is this post misogynistic?} 
\end{figure}
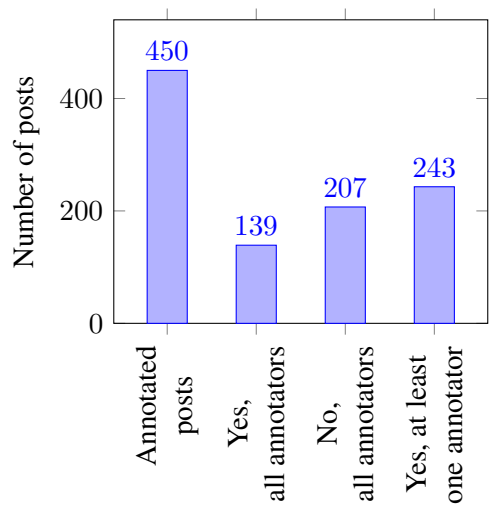

\textbf{Hate speech}
Our experts were asked to perform binary classification of whether the post they were reading contained hate speech or not. This gave the experts the chance to mark any type of hate speech or hateful behavior.

\textbf{Misogyny}
The second task was the binary classification of misogyny based on the instructions and the definition of misogyny provided in the annotation guidelines.

\textbf{Category}
Once a post was classified as misogynistic, our experts were requested to choose a category label. Our taxonomy of misogyny consists of the following categories: stereotype, erasure and minimization, violence against women, sexualisation and objectification, anti-feminism and denial of sexualisation. The experts could only choose one category and could not choose a subcategory outside of the ones presented.

\textbf{Severity}
Experts hold that misogyny exists in a spectrum and it depends on individual perception. To portray this, we asked them to give a score ranging from 1 to 10, where 1 is the least misogynistic. Although one would assume that, for example, a post portraying violence would have a high score, we did not give them any specific guidelines they had to follow to assign these scores and asked them to trust their own judgement.

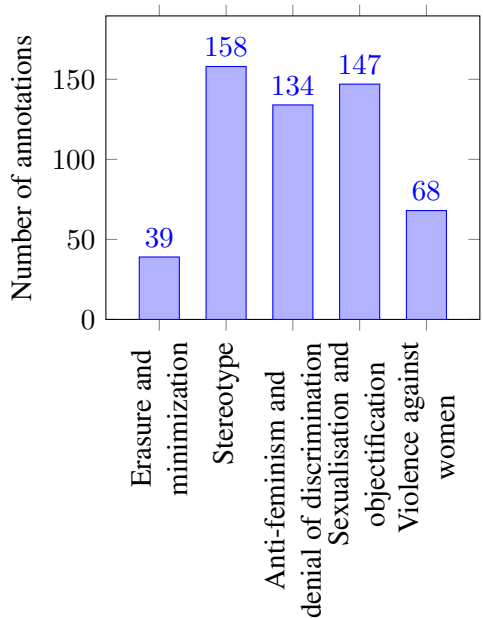
\begin{figure}
    \centering
    \begin{tikzpicture}
        \begin{axis}[
                ybar,
                symbolic x coords={A, B, C, D, E},
                xtick=data,
                ylabel={Number of annotations},
                xticklabel style={rotate=90, anchor=east, align=center},
                xticklabels={
                    {Erasure and\\minimization}, 
                    {Stereotype}, 
                    {Anti-feminism and\\denial of discrimination}, 
                    {Sexualisation and\\objectification},
                    {Violence against\\women}
                },
                ymin=0,
                bar width=15pt,
                nodes near coords,
                enlarge x limits=0.2,
                enlarge y limits={upper, value=0.2}
                ]
        \addplot coordinates {(A, 39)
                             (B, 158) 
                             (C, 134)
                             (D, 147)
                             (E, 68)
                             };
        \end{axis}
    \end{tikzpicture}
    \caption{Which category of misogyny does this example belong to?} 
\end{figure}

\section{Results}

This section provides an overview of the annotation results.

\textbf{Hate speech}
Each of the 450 posts was annotated by two to four experts and in almost two-thirds of the cases all experts agreed on whether hate speech was present.
In the 334 cases where all experts agreed, slightly more posts were annotated as containing hate speech compared to not containing hate speech but the difference was marginal. However, there was a large difference between the number of posts that were annotated as hate speech by all experts (172) and posts that were considered to be hate speech by at least one expert (288).

\textbf{Misogyny}
Figure 1 shows the annotation results for the misogyny classification task. 
A negative label in the previous task is considered to be a negative label in the misogyny classification task as well, in which case, all 450 posts were again annotated by two to four experts. In slightly more than two-thirds of the examples, all experts agreed on the label. However, in this case, there was a larger imbalance between the two possible labels: 207 posts were considered to be non-misogynistic and 139 misogynistic by all experts. The number of posts considered to be misogynistic by at least one expert was however larger at 243 posts.

\textbf{Category}
There was more disagreement in choosing the category of misogyny. There 
were between zero and four category annotations for each of the 450 posts in the dataset. Out of the 185 posts in the final dataset with more than one category annotation, in 93 cases all experts chose the same category. Figure 2 gives an overview of all 546 category annotations in the dataset, comparing the number of times each of the five possible categories was chosen.

\textbf{Severity}
The last annotation task asked the experts to estimate the severity of the misogyny in the post. Similarly to the previous task, 185 posts had at least two severity annotations and a closer analysis of those revealed that although the experts seldom selected the same rating, in 91\% of cases the difference between the minimum and the maximum rating was not larger than 3.

\section{Discussion and Conclusion}

This project’s primary contribution lies in its interdisciplinary approach to misogyny detection in Swedish rather than the dataset itself, which remains small. Collaborating with experts from diverse fields, we developed an annotation process that captures the complexity of misogyny as it manifests in Swedish online discourse. This experiment provides a valuable framework for future studies focused on bias detection in under-resourced languages.

In the context of misogyny detection, defining what constitutes misogynistic language is inherently challenging. Attempting to capture a wider range of potentially harmful expressions risks being too broad, while using a stricter approach might fail to recognize subtler forms of misogyny. The challenge lies in determining who defines misogyny, as cultural, linguistic and societal factors have an influence over the definition, making it a complex decision.

The feedback from the experts highlighted the need for clearer operational definitions and stronger contextual support. Misogyny detection, particularly in a complex environment like Flashback, would benefit from additional discussion on cultural nuances and interdisciplinary perspectives. Additionally, better alignment between academic rigor and practical applicability is crucial to ensuring that interdisciplinary projects like this one fully realize their potential. We also took into account the experts' perspective in section \ref{Limitations and Future Work}.

In conclusion, while the dataset is limited, the interdisciplinary approach and methodology offer a valuable starting point for future research. Refining the annotation process and expanding the dataset could further improve the effectiveness of misogyny detection tools, especially for lower-resourced languages like Swedish.

\section{Limitations and Future Work}
\label{Limitations and Future Work}
This project has several limitations that provide avenues for future work.

\textbf{Dataset Size and Diversity}
The current dataset, while robustly annotated, is relatively small, limiting its capacity to capture the full spectrum of misogynistic expressions within the Swedish online discourse. The limited number of examples might not adequately represent less overt forms of misogyny, which are increasingly prevalent and harmful. Future work should focus on expanding the dataset to include a larger variety of sources.

\textbf{Keyword Selection Bias}
The reliance on predefined keywords to scrape forum posts inherently introduces selection bias, primarily focusing on explicit forms of misogyny. This method may overlook subtle or emergent forms of misogynistic language that do not necessarily conform to expected patterns. Future iterations of this project should aim to refine the keyword selection process. Additionally, incorporating machine learning techniques to identify potential posts could reduce bias introduced by keyword dependency.

\textbf{Decontextualisation}
A key challenge was annotating decontextualised posts, which made it difficult to detect subtle misogyny. Without context, the experts had to rely on isolated phrases, often missing nuances that could clarify intent or severity. Providing more context in future datasets would enhance accuracy.

\textbf{Consensus Building} 
Disagreements among the experts highlighted the subjective nature of misogyny detection and the challenges in classifying complex human behaviors and attitudes. While we utilized workshops to align annotator perspectives, a more systematic approach to handling disagreement could enhance the consistency and reliability of annotations. Future work could include developing detailed guidelines based on the initial rounds of annotation to standardize responses and improve inter-annotator reliability. Implementing an adjudication process where the experts discuss and resolve disagreements before finalizing annotations could also be beneficial.

\section*{Acknowledgments}

This work is a result of the “Interdisciplinary Expert Pool for NLU” project funded by Vinnova (Sweden’s innovation agency) under grant 2022-02870.

\subsection*{Experts Involved}
\begin{raggedright}
\begin{itemize}[noitemsep, topsep=0pt] 
    \item Annika Raapke, Researcher at Uppsala University, Department of History
    \item Eric Orlowski, Sociocultural Anthropologist, Research Fellow (AI Governance), AI Singapore
    \item Michał Dzieliński, Assistant Professor at Stockholm Business School, International Finance
    \item Maria Jacobson, Anti-Discrimination Agency West Sweden
    \item Astrid Carsbrin, Swedish Women's Lobby
    \item Cia Bohlin, The Swedish Internet Foundation
    \item Richard Brattlund, The Swedish Internet Foundation
\end{itemize}

\subsection*{Special Thanks}
Special thanks to:
\begin{itemize}[noitemsep, topsep=0pt]
    \item Francisca Hoyer at AI Sweden for making the Interdisciplinary Expert Pool possible from the start
    \item Magnus Sahlgren at AI Sweden for guidance
    \item Allison Cohen at MILA AI for Humanity for participation and support during the experiment
\end{itemize}
\end{raggedright}

BiaSWE's multi-disciplinary engagement process was, in part, inspired by the Biasly project from Mila - Quebec AI Institute.

This work was partially supported by the Wallenberg AI, Autonomous Systems and Software Program (WASP) funded by the Knut and Alice Wallenberg Foundation.

\bibliographystyle{acl_natbib}
\sloppy
\bibliography{nodalida2025}

\end{document}